\documentclass[letterpaper, 10 pt, conference]{ieeeconf}  
\usepackage{hhline}
\usepackage{graphicx}
\usepackage{subcaption}
\usepackage{multirow}
\usepackage{soul,color}
\usepackage{amsfonts}
\usepackage{float}
\usepackage{amsmath}
\usepackage{algorithm}
\usepackage{algorithmic}
\usepackage{boldline} 

\newcommand{\bigO}{\mathcal{O}}

\usepackage{kotex}
\usepackage{kotex-logo}
\IEEEoverridecommandlockouts                              

\title{\LARGE \bf
Memory-based gaze prediction in deep imitation learning for robot manipulation
}
\author{Heecheol Kim$^{1}$$^{,}$$^{2}$, Yoshiyuki Ohmura$^{1}$, Yasuo Kuniyoshi$^{1}$%
\thanks{$^{1}$ Laboratory for Intelligent Systems and Informatics, Graduate School of Information Science and Technology, The University of Tokyo, 7-3-1 Hongo, Bunkyo-ku, Tokyo, Japan (e-mail: \{h-kim, ohmura, kuniyosh\}@isi.imi.i.u-tokyo.ac.jp, Fax: +81-3-5841-6314) }%
\thanks{$^{2}$ Corresponding author}
\thanks{This paper is supported in part by the Department of Social Cooperation Program ``Intelligent Mobility Society Design'', funded by Toyota Central R\&D Labs., Inc.,  of the Next Generation AI Research Center, The University of Tokyo.}
}
\begin{document}
\maketitle
\thispagestyle{empty}
\pagestyle{empty}

\begin{abstract}
Deep imitation learning is a promising approach that does not require hard-coded control rules in autonomous robot manipulation. The current applications of deep imitation learning to robot manipulation have been limited to reactive control based on the states at the current time step. However, future robots will also be required to solve tasks utilizing their memory obtained by experience in complicated environments (e.g., when the robot is asked to find a previously used object on a shelf). In such a situation, simple deep imitation learning may fail because of distractions caused by complicated environments. We propose that gaze prediction from sequential visual input enables the robot to perform a  manipulation task that requires memory. The proposed algorithm uses a Transformer-based self-attention architecture for the gaze estimation based on sequential data to implement memory.
The proposed method was evaluated with a real robot multi-object manipulation task that requires memory of the previous states.

\end{abstract}

\providecommand{\keywords}[1]{\textbf{\textit{Index terms---}} #1}
 

\section{Introduction}

\begin{figure*}
  \centering
  \vspace{0.0in}
  \includegraphics[width=0.85\linewidth]{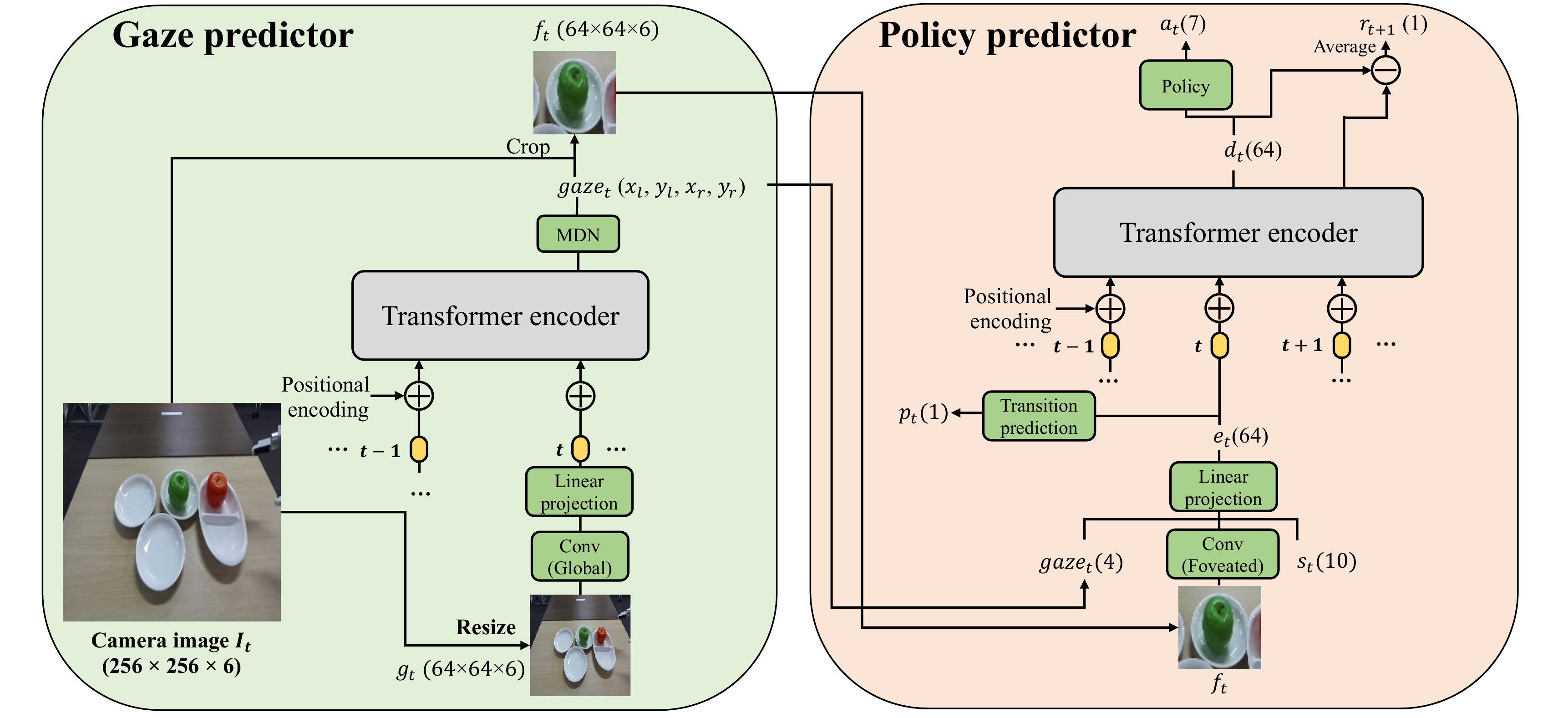}
  \captionsetup{justification=centering}
  \caption{Proposed deep imitation learning architecture for sequential data while training. During execution, $r_{t+1}$ is computed at the next time step $t+1$ (see Algorithm \ref{alg:algo}); (n) indicates dimensions.}
  \label{fig:entire_architecture}
 \end{figure*}
 
In robot manipulation, the application of deep imitation learning, which trains a deep neural network controller from the expert's demonstration data, has made several dexterous robot manipulation tasks possible without the need for pre-defined robot control rules (e.g., \cite{zhang2018deep,kim2021gaze,kim2021transformer}).

The current deep imitation learning architectures for robot manipulation infer a reactive action to the current state. This limits the deep imitation learning-based robot manipulation to the real-world applications, where a robot may be required to utilize memory to accomplish a task. For example, when the robot is asked to find an object in a closed cupboard, it has to recall where the object was placed. 
Especially, those real-world applications will frequently be involved with the complicated environment with multiple, cluttered objects. The simple end-to-end recurrent deep imitation learning which inputs the sequence of the entire image fails in such a complicated multi-object environment because of distractions caused by multiple task-unrelated objects \cite{kim2020using} and the distractions inherent in the expanded state space.

The implementation of human-like gaze to robot vision for deep imitation learning was first studied in our previous work and found to be effective in the manipulation of multiple objects \cite{kim2020using} or dexterous manipulation of a very small object \cite{kim2021gaze}. In these studies, the expert's eye gaze while operating the robot to generate the demonstration data was collected by an eye-tracker, and the mapping of the robot vision to the eye gaze was trained by a neural network that estimates the probability distribution of the given data. 
The gaze-based method has a few advantages over pre-trained object detectors (e.g., \cite{xiang2018posecnn,tremblay2018corl:dope}). First, it requires no annotation but human gaze, which can be naturally collected during the demonstration. Second, the gaze signal contains information about intention; therefore, the target object can be distinguished from background objects. 
However, these methods predict the probability distribution of gaze from current visual input only; therefore, they lack the capability to consider memory.

We propose that a sequential data-based gaze control can be used to achieve memory-based robot manipulation. Studies in psychology suggest humans use memory representation, as well as visual input for gaze control \cite{li2018memory,hayhoe2018control}. 
Therefore, the memory-based gaze generation system can provide sufficient information needed for the manipulation.
For example, when a human recalls the location of an object in the closed cupboard, the human first gazes at the remembered location of the object and then attempts to manipulate it.
This method controls gaze using not only the current visual input but also previous vision data; therefore, it enables the robot to determine the correct location, which is closely related to the current robot manipulation policy and can only be inferred from the data of the previous time step.

We compared gaze predictors using two different types of neural network architecture that process sequential visual input.
The first method for sequential data processing is the recurrent neural network (RNN) \cite{hochreiter1997long,cho2014learning}. An RNN processes sequential data input recursively and updates its internal state. This method has been widely used in domains with sequential data (e.g., speech recognition \cite{graves2013hybrid} and machine translation \cite{sutskever2014sequence}).
The second method, self-attention was proposed relatively recently first for natural language processing \cite{vaswani2017attention}. The Transformer, a variant of a neural network that implements self-attention, analyzes relationships between data embeddings so that it can pay attention to important embeddings.

The proposed sequential robot imitation learning architecture was tested on a multi-object manipulation task that requires memory of the previous time steps.

\section{Related work}

\subsection{Gaze-based deep imitation learning}
Gaze-based deep imitation learning can generate task-related spatial attention by using the gaze signal which can be naturally acquired during teleoperation. For example, \cite{zhang2018agil} trained deep neural network using human demonstration with gaze while playing Atari video game. Our previous research investigated gaze-based deep imitation learning for robot manipulation to suppress visual distractions \cite{kim2020using}. \cite{kim2021gaze} proposed a gaze-based dual resolution system for separate fast-speed and slow-speed action control applied to precise manipulation tasks.

The previous gaze-based deep imitation learning methods considered reactive gaze control from current visual input. However, \cite{li2018memory} and \cite{hayhoe2018control} suggested that the human's eye movement also depends on memory as well as the current visual information. Therefore, memory dependent gaze system for deep imitation learning may better reconstruct the human's gaze model and provides benefits on automated manipulation tasks.

\subsection{Simulating memory in robots}
Previous researches have implemented memory in robots. For example, \cite{phillips2005biologically} studied a neural network-based temporal difference (TD) learning for a delayed saccade task. However, this study did not evaluate if the gaze saccade can improve the motor command of the robot. 
\cite{gordon2006system} proposed behavior learning with working memory for robot manipulation tasks. Nevertheless, this requires a pre-trained vision perception system for object detection. \cite{hussein2018deep} tested an RNN-based imitation learning for Robocup \cite{kitano1997robocup}, but this research did not consider visual attention. Unlike the studies above, our work validates that gaze control improves memory-based manipulation.

\section{Method}
\subsection{Gaze prediction for sequential robot data}

In the gaze-based deep imitation learning framework \cite{kim2020using}, a human expert teleoperates a UR5 (Universal Robots) and its gripper. A stereo image captured from a stereo camera mounted on the robot is provided to the human operator through a head-mounted display. During operation, a Tobii eye-tracker on the head-mounted display captures eye movement and calculates the two-dimensional gaze position.

In the previous method, which does not consider a temporal sequence as input data, a neural network is trained to predict the current gaze position $gaze_t$ from the current stereo image $g_t$:
\begin{equation}
\begin{aligned}
\label{eq:gaze_current}
gaze_t \sim p_{\theta}(g_t),
\end{aligned}
\end{equation}
where $\theta$ indicates the neural gaze predictor. Because this paper extends the gaze predictor into sequential data, the gaze is computed by following:
\begin{equation}
\begin{aligned}
\label{eq:gaze_sequential}
gaze_t \sim p_{\theta}(g_0, g_1, ..., g_t).
\end{aligned}
\end{equation}

Because the gaze may change drastically in a similar situation \cite{kim2020using}, a mixture density network (MDN) \cite{bishop1994mixture}, which computes the probability distribution of given data by fitting neural network output to a Gaussian mixture model (GMM), is used to infer the probability distribution of the gaze. The MDN outputs the expected gaze position $\mu_t^i$, covariance matrix $\mathbf{\Sigma}_t^i=\begin{bmatrix} {{\sigma_x}_t^i}^2 & \rho_t^i {\sigma_x}_t^i {\sigma_y}_t^i \\ \rho_t^i {\sigma_x}_t^i {\sigma_y}_t^i & {{\sigma_y}_t^i}^2 \end{bmatrix}$, and the weight of each Gaussian $w^i_t$. Then, this probability estimation is optimized to true estimated human gaze $e_t$ by negative log-likelihood loss \cite{kim2020using,kim2021transformer}: 

\begin{equation}
\label{eq:loss_gaze}
\mathcal{L}_{gaze} = -log \big{(}\sum_{i=1}^{N'} w_t^i \mathcal{N} (h_t;\mu_t^i,\mathbf{\Sigma}_t^i)\big{).}
\end{equation}
We used $N'=8$ Gaussian distributions to follow the MDN architecture used in \cite{kim2021gaze}.

This proposed gaze predictor architecture is similar to the architecture described in \cite{bazzani2016recurrent} which proposed a long short-term memory (LSTM) based recurrent MDN architecture for the saliency prediction in video data.
However, our study aims to show that when the manipulation policy relies on a specific point of memory, its efficient reasoning is possible with gaze prediction using memory, whereas the aim of \cite{bazzani2016recurrent} is to predict the probability of saliency in a video clip.

\subsection{Transformer-based sequential robot data processing}
In the proposed architecture (Fig. \ref{fig:entire_architecture}), the Transformer is used to process sequential data captured from the robot's sensors. During training, each time step $t \in [0, N]$ in the sequence with length $N$ is associated with a camera image $I_t$ and robot kinematics states $s_t$, which is represented by a 3-dimensional position, a 6-dimensional posture represented by a combination of the cosine and sine of each Euler angle, and the gripper angle.

First, the gaze predictor extracts the gaze position from the sequence of the camera input sequence. The left/right $256 \times 256$ RGB images captured from the stereo camera resized into $64 \times 64$ images to form global image $g_t$. Then, $g_{0..N}$ is used to predict gaze $gaze_{0..N}$. Each global image $g_t$ is processed using a five-layer convolutional neural network (Conv(global)) to extract visual features and transformed into 16 spatial feature points using spatial softmax \cite{finn2016deep}. 
These spatial features are processed by the Transformer encoder with a mask that masking out attention to the future so that the feature embeddings only depend on past embeddings. 
The output of the Transformer encoder at the time step $t$ is processed by an MDN with a multi-layer perceptron (MLP) with two hidden layers of size 200. Eight Gaussians are used to fit the probability distribution of the gaze. The two-dimensional gaze position with maximum probability is sampled and used to crop a $64 \times 64$ foveated image from the $256 \times 256$ camera image $I_t$.

The policy predictor also uses sequential data for consistency with the gaze predictor.
The policy predictor uses the sequence of gaze position inferred from the gaze predictor, the foveated image, and the 10-dimensional kinematics states of the robot to predict the action defined by the difference between the next target state and the current state: $a_t = s_{t+1} - s_t$. Each foveated image $g_t$ is processed with series of convolutional layers (Conv(foveated)) and its number of parameters is reduced by a global average pooling (GAP) layer \cite{lin2013network}, which averages the feature map into one value. This image representation is concatenated with gaze position $gaze_t$ and robot state $s_t$, and then passed to linear projection to form the embedding vector $e_t$. The sequence of this concatenated embedding $e_t$ is sent to the Transformer encoder, which uses masked multi-head attention so that output action $a_t$ does not depend on the future state. The output representation $d_t$  is sent to a two-layered MLP with a hidden-layer size of 200 to compute the final action output $a_t$. All convolutional layers use a kernel size of three and padding of one throughout this research. We used the $\ell_2$ loss for the policy predictor and MDN loss for the gaze predictor. We refer readers to \cite{kim2021transformer} for more detailed information on convolutional layers.

\subsection{Prediction of the representation transition}
\begin{algorithm}[tb]
\caption{Proposed algorithm for sequential robot data during execution.}
\label{alg:algo}
\textbf{Parameter}: Policy predictor $\pi$, gaze predictor $\theta$, transition threshold $C = 0.9$,  foveated image sequence $L_f$, global image sequence $L_g$, and robot state sequence $L_s$. \\
\begin{algorithmic}[1] 
\STATE $t \gets 0$, $succeed \gets $ false
\STATE $f_1 \gets \bf{0}$ \COMMENT{initialize foveated image with zeros}

\WHILE{$\neg$ $succeed$}
    \STATE $t \gets t + 1$
    \STATE append $f_{t-1}$ to $L_f$
    \STATE $I_t, s_t \gets$ robot sensor image $\&$ kinematics state
    \STATE $g_t \gets resize(I_t, (64, 64))$
    \STATE $gaze_t \gets \theta (L_g)$
    \STATE $f_t \gets crop(I_t, gaze_t)$
    \STATE append $g_t$ to $L_g$, $s_t$ and $gaze_t$ to $L_s$
    \STATE $a_t, r_t, p_t \gets \pi(L_f, L_s)$ \COMMENT{predict action $a_t$, representation transition $r_t$, prediction of the representation transition $p_t$}
    \IF{$r_t < C p_{t-1}$}
        \STATE delete $g_t$ from $L_g$, $s_t$ and $gaze_t$ from $L_s$, and $f_{t-1}$ from $L_f$
        \STATE $t \gets t-1$
    \ENDIF
    \STATE Execute $a_t$ on the robot 
    \STATE Manually decide $succeed$
\ENDWHILE
\end{algorithmic}
\end{algorithm}

\begin{table}
\centering
\begin{tabular}{lccc}
\hlineB{2}
Criteria  &  Non-sequential     &    LSTM        & Transformer                   \\ \hline\hline
(a) Successful gaze & 50.6\% & 100.0\% & 100.0\% \\ \hline
(b) Euclidean distance & 16.8 (21.6) &14.9 (19.3) & 14.8 (18.2) \\ \hline
\hlineB{2}
\end{tabular}
\caption{(a) Rates of the successful gazes at the green apple's previous position. (b) Mean (SD) Euclidean distance between human gaze and prediction (in pixels).}
\label{tab:gaze_accuracy}
\end{table}
 
\begin{figure}
  \centering
  \vspace{0.0in}
  \includegraphics[width=0.8\linewidth]{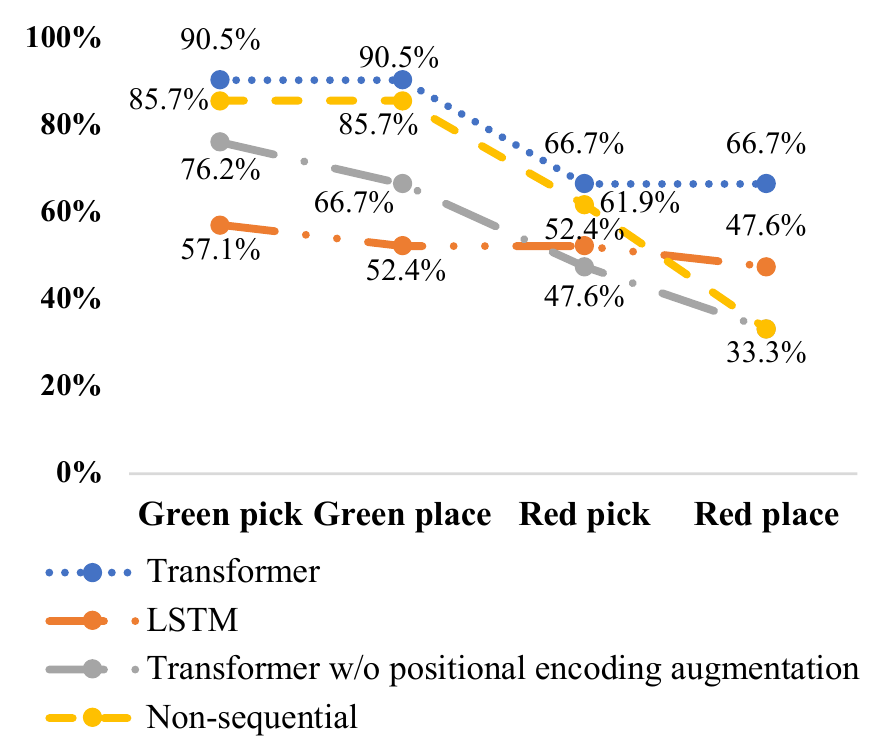}
  \captionsetup{justification=centering}
  \caption{Task success rate comparison (21 trials).}
  \label{fig:accuracy}
 \end{figure}

\begin{figure*}
  \centering
  \vspace{0.0in}
  \begin{subfigure}[t]{.5\linewidth}
    \captionsetup{width=1.\linewidth}
    \captionsetup{justification=centering}
    \includegraphics[width=0.98\linewidth]{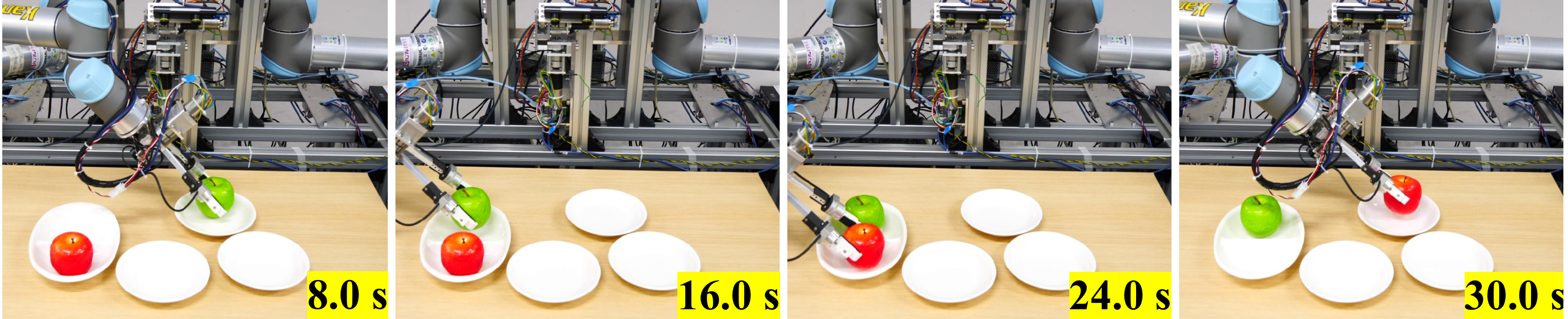}
    \caption{Example 1}
    \label{fig:transformer_success0}
  \end{subfigure}%
  \begin{subfigure}[t]{.5\linewidth}
    \captionsetup{width=1.\linewidth}
    \captionsetup{justification=centering}
    \includegraphics[width=0.98\linewidth]{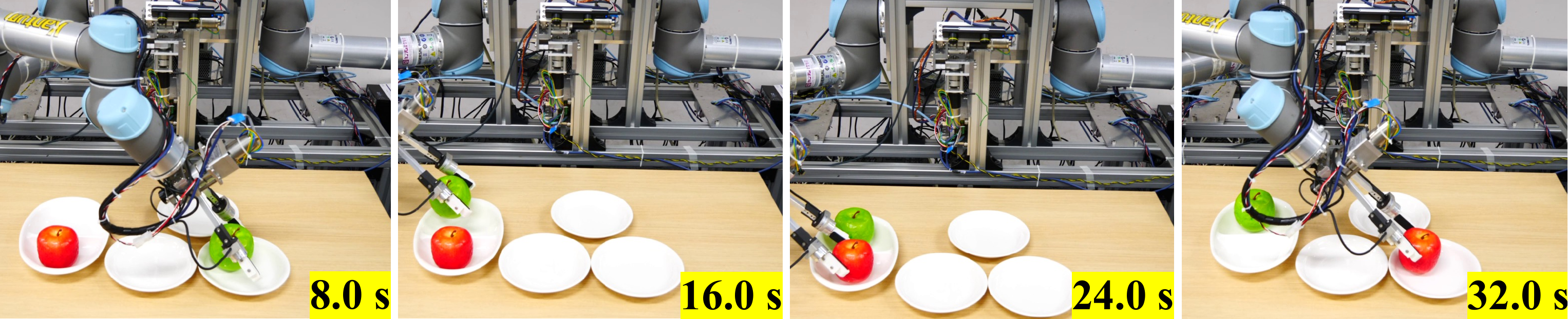}
    \caption{Example 2}
    \label{fig:transformer_success1}
  \end{subfigure}%
      \vskip\baselineskip

  \begin{subfigure}[t]{.99\linewidth}
    \captionsetup{width=1.\linewidth}
    \captionsetup{justification=centering}
    \includegraphics[width=0.98\linewidth]{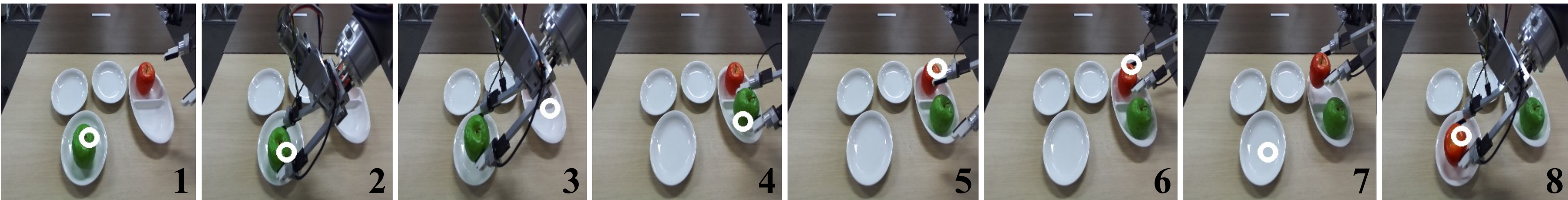}
    \caption{Robot's view of Example 1 with the predicted gaze position (left image).}
    \label{fig:transformer_success1_gaze}
  \end{subfigure}%

  \captionsetup{justification=centering}
  \caption{Examples of successful manipulation trials using the Transformer-based method. The robot was able to gaze at the plate where the green apple had been located and place the red apple on the correct plate.}
  \label{fig:transformer_successful}
 \end{figure*}
 
\begin{figure*}
  \centering
  \vspace{0.0in}
  \begin{subfigure}[t]{.5\linewidth}
    \captionsetup{width=1.\linewidth}
    \captionsetup{justification=centering}
    \includegraphics[width=0.98\linewidth]{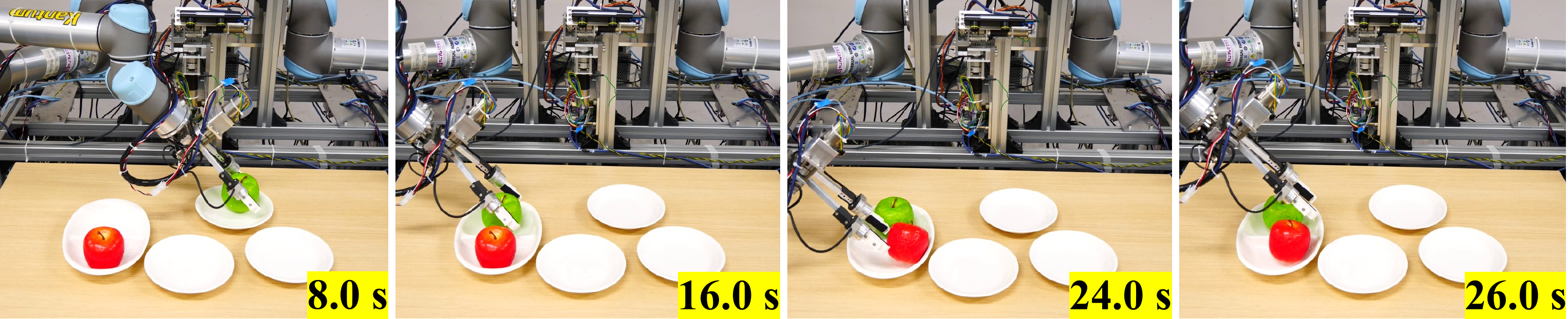}
    \caption{Example 1: failure to pick the red apple.}
    \label{fig:ep_failure0}
  \end{subfigure}%
  \begin{subfigure}[t]{.5\linewidth}
    \captionsetup{width=1.\linewidth}
    \captionsetup{justification=centering}
    \includegraphics[width=0.98\linewidth]{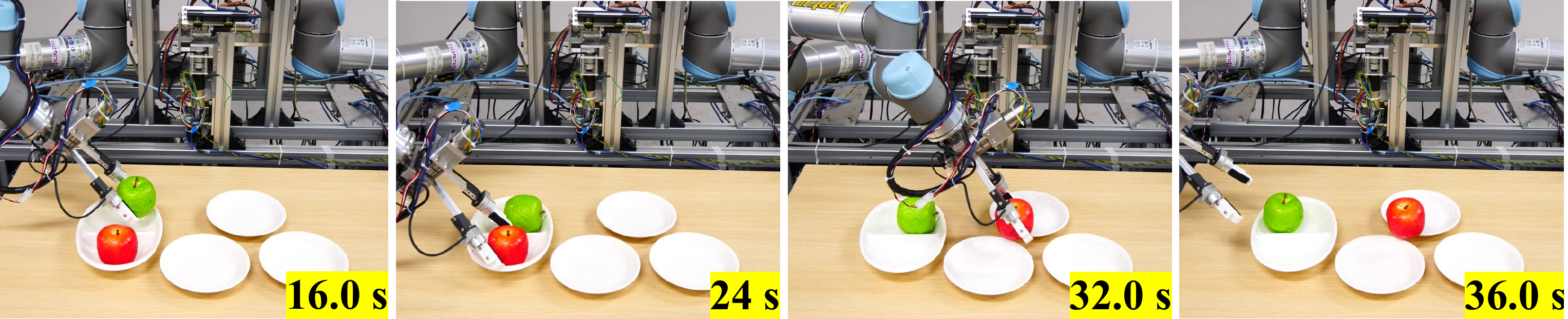}
    \caption{Example 2: failure to find the green apple's previous location.}
    \label{fig:ep_failure1}
  \end{subfigure}%
      \vskip\baselineskip

  \begin{subfigure}[t]{.99\linewidth}
    \captionsetup{width=1.\linewidth}
    \captionsetup{justification=centering}
    \includegraphics[width=0.98\linewidth]{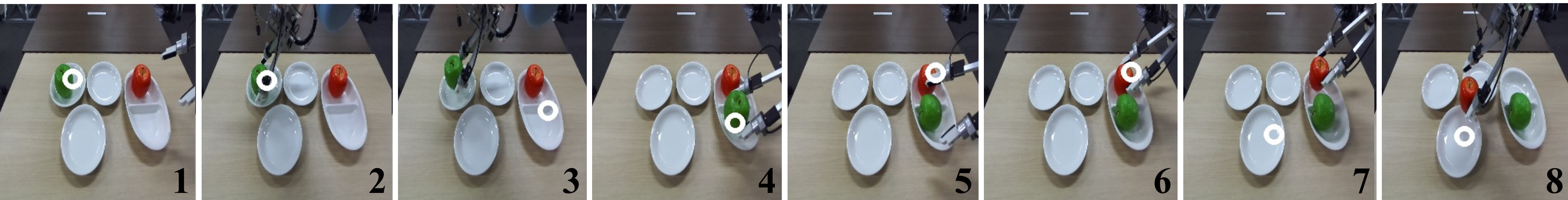}
    \caption{Robot's view of Example 2 with the predicted gaze (left image). In image 5, the robot gazed at a different plate that had not held the green apple.}
    \label{fig:ep_failure1_gaze}
  \end{subfigure}%

  \captionsetup{justification=centering}
  \caption{Examples of failures of the non-sequential model.}
  \label{fig:ep_failure}
 \end{figure*}
 
During the execution of the trained sequential model on a real robot, we observed that changes in the robot state became smaller while grasping, which resulted in unexpected stationary state points.
This causes a discrepancy between the training dataset and the observed sequence over time. The sequential discrepancy accumulates small mistakes in the action execution that may lead to failure \cite{ross2011reduction}. 

The transition prediction method, which excludes stationary points, is hence proposed to prevent such stationary points from affecting the neural network output.
The idea is that if the predicted action $a_t$ has not sufficiently changed a representation at the next time step $d_{t+1}$, the current time step is excluded from the input sequence so that it cannot be used in further inference. Specifically, a representation transition $r_{t+1}$ at time step $t+1$ is computed from two adjacent representations passed through the Transformer encoder as follows: 
\begin{equation}
\begin{aligned}
\label{eq:representation_transition}
r_{t+1} = \mathbb{E} (d_{t+1} - d_t).
\end{aligned}
\end{equation}

Then, the embedding $e_t$ at time step $t$, which consists of the concatenated and linear projected feature of $gaze_t$, a representation of the foveated image $f_t$ after convolution, and the robot's kinematic states $s_t$, is processed with two MLP layers to output a prediction $p_t$. $p_t$ is the prediction of $r_{t+1}$ which is the representation transition at the next time step $t+1$. During execution, if $r_{t+1}$ is smaller than $C p_t$, where $C$ represents the transition threshold, the data at time step $t$ is excluded from the sequence to suppress the stationary points (see Algorithm \ref{alg:algo}).

We found the policy prediction using this exclusion method yielded a higher success rate than reactive policy prediction. However, because the focus of this paper is memory-based gaze prediction for robot manipulation, a detailed analysis of the exclusion method is left for future work.

\subsection{Positional encoding augmentation}
Positional encoding is widely used in Transformer architectures to provide the positional information of each embedding \cite{vaswani2017attention}. 
Sinusoidal positional encoding is one of the most widely used approaches \cite{vaswani2017attention}, and is expressed as follows:
\begin{equation}
\begin{aligned}
PE_{t, 2i} = sin(\frac{t}{W^{\frac{2i}{ch}}}), 
PE_{t, 2i+1} = cos(\frac{t}{W^{\frac{2i}{ch}}}),
\label{eq:sinusoid}
\end{aligned}
\end{equation}
where $t \in [0, N]$ refers to the position of the embedding, $ch$ indicates the dimensionality of the embedding, and $i$ indicates each dimension.

To reduce the effect of the distortion in time series caused by excluded stationary points, the scaling factor $W$ of the sinusoidal positional encoding is randomly augmented by a value between $2,000$ -- $20,000$ in every minibatch during training so that the positional encoding is robust against the distortion. The range of the augmentation values is selected to cover the range include $10,000$, which is the commonly used value of $W$\cite{vaswani2017attention}. 
During the execution, $W$ was fixed to the value of $10,000$. Because positional information only depends on positional encoding in the Transformer, this augmentation can simulate a training dataset with different time series intervals.

\section{Experiments}

\subsection{Task setup}
We designed the \textit{Replace} task to be a task that requires the memory of previous states. 
In this task, three plates are located on a table and a green toy apple is randomly placed on one of the plates. The robot is required to grasp this green apple and place it on the target plate (which is placed to the right), then pick up a red toy apple (which has already been placed on the target plate) and place it on the plate that originally held the green apple.
Even though this task is simple for a human, the robot must memorize the initial location of the green apple and use that memory to control the current manipulation. In addition, because this task requires the multi-object manipulation of two apples, it requires the use of gaze-based deep imitation learning \cite{kim2020using}. 
The total dataset was divided into a $90\%$ training dataset and $10\%$ validation dataset. The training dataset consists of $824$ demonstrations with a total demonstration time of $186.5$ minutes; therefore, the average (SD) episode length is $13.6 (1.96)$ seconds.

\subsection{Baseline architectures}
We used two baseline architectures for comparison with the proposed Transformer-based imitation learning for temporal robot data. First, a non-sequential baseline that does not utilize any previous time steps was investigated to evaluate the necessity of sequential data in the memory-based gaze prediction. In this architecture, the Transformer encoder layer of the gaze predictor is replaced with one fully connected layer so that the network cannot consider previous time steps. The aim of the second LSTM baseline architecture is to compare the self-attention and RNNs. In this architecture, the Transformer encoder layer is replaced with one LSTM layer and the transition prediction is used. Because \textit{Replace} is a multi-object manipulation task, a vanilla deep imitation learning without gaze fails \cite{kim2020using}. Therefore, deep imitation learning without gaze is excluded from this experiment. 

\subsection{Performance evaluation}

 \begin{figure}
  \centering
  \vspace{0.0in}
  \includegraphics[width=0.98\linewidth]{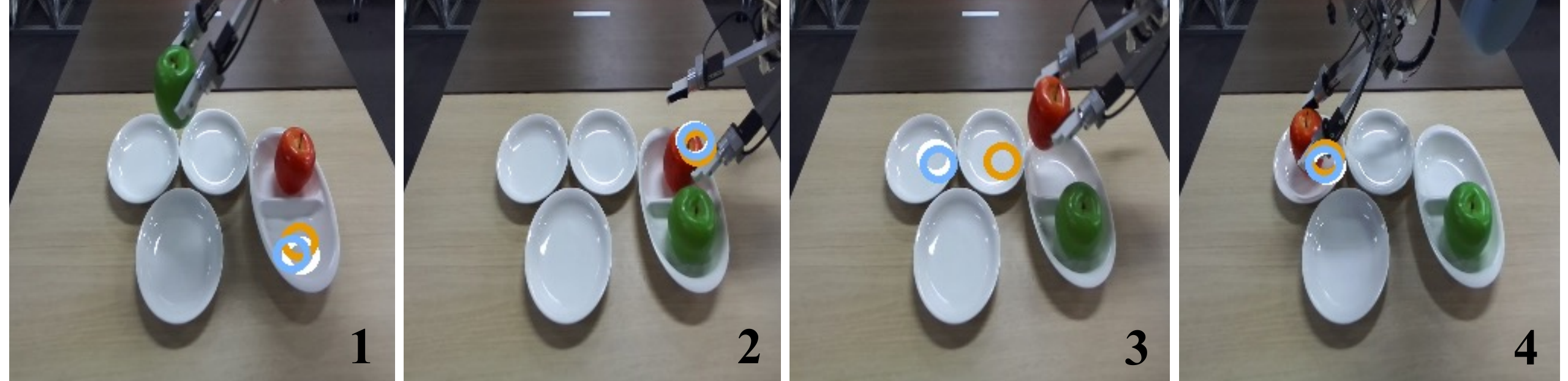}
  \captionsetup{justification=centering}
  \caption{Example gaze behaviors on a validation episode (Human gaze: white, Transformer: blue, Non-sequential: orange).}
  \label{fig:gaze_comp}
 \end{figure}

\begin{figure*}
  \centering
  \vspace{0.0in}
  \begin{subfigure}[t]{.25\linewidth}
    \captionsetup{width=1.\linewidth}
    \captionsetup{justification=centering}
    \includegraphics[width=0.98\linewidth]{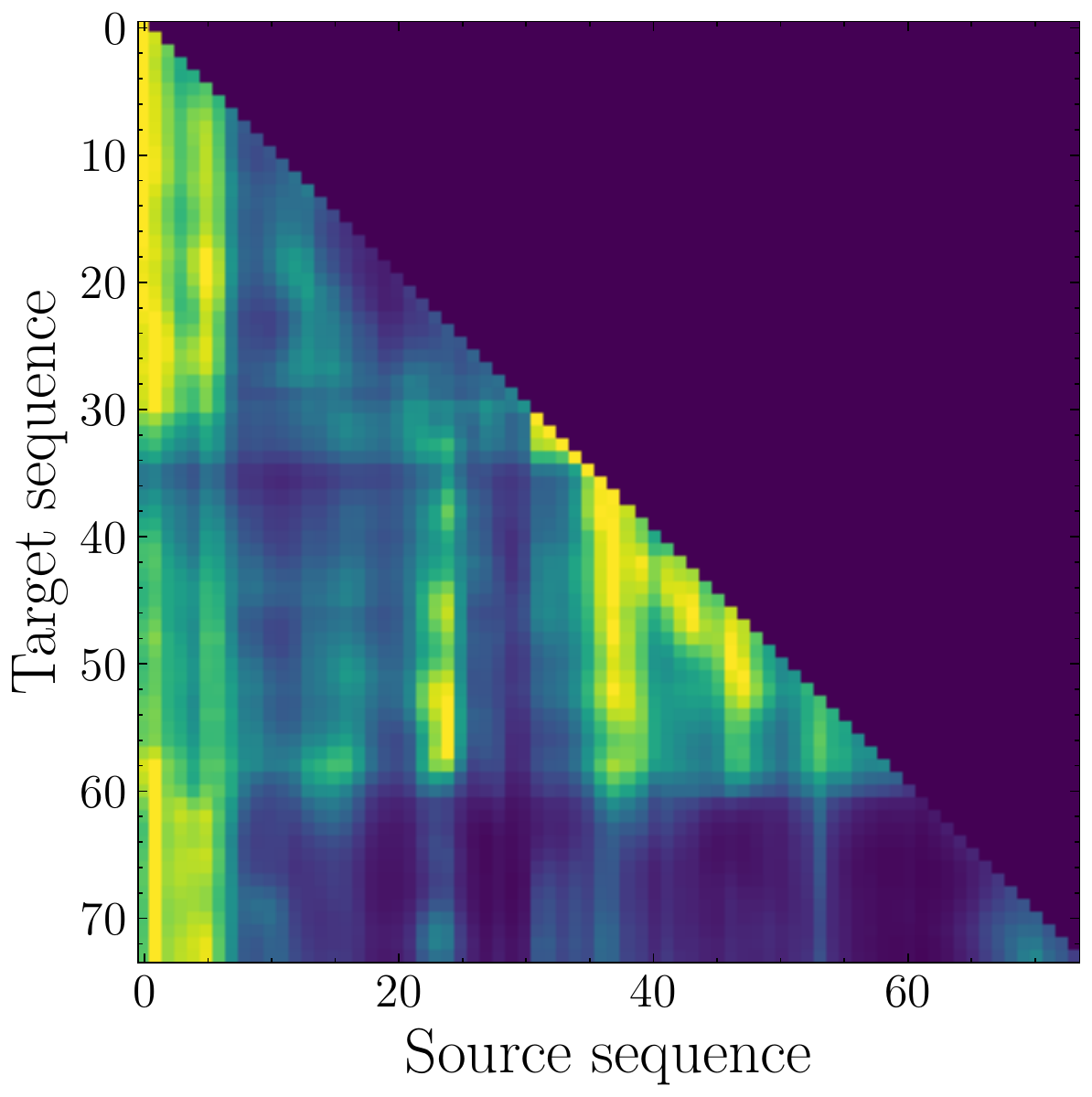}
    \caption{Example of sequential attention map.}
    \label{fig:attention_map}
  \end{subfigure}%
  \begin{subfigure}[t]{.48\linewidth}
    \captionsetup{width=.9\linewidth}
    \captionsetup{justification=centering}
    \includegraphics[width=0.98\linewidth]{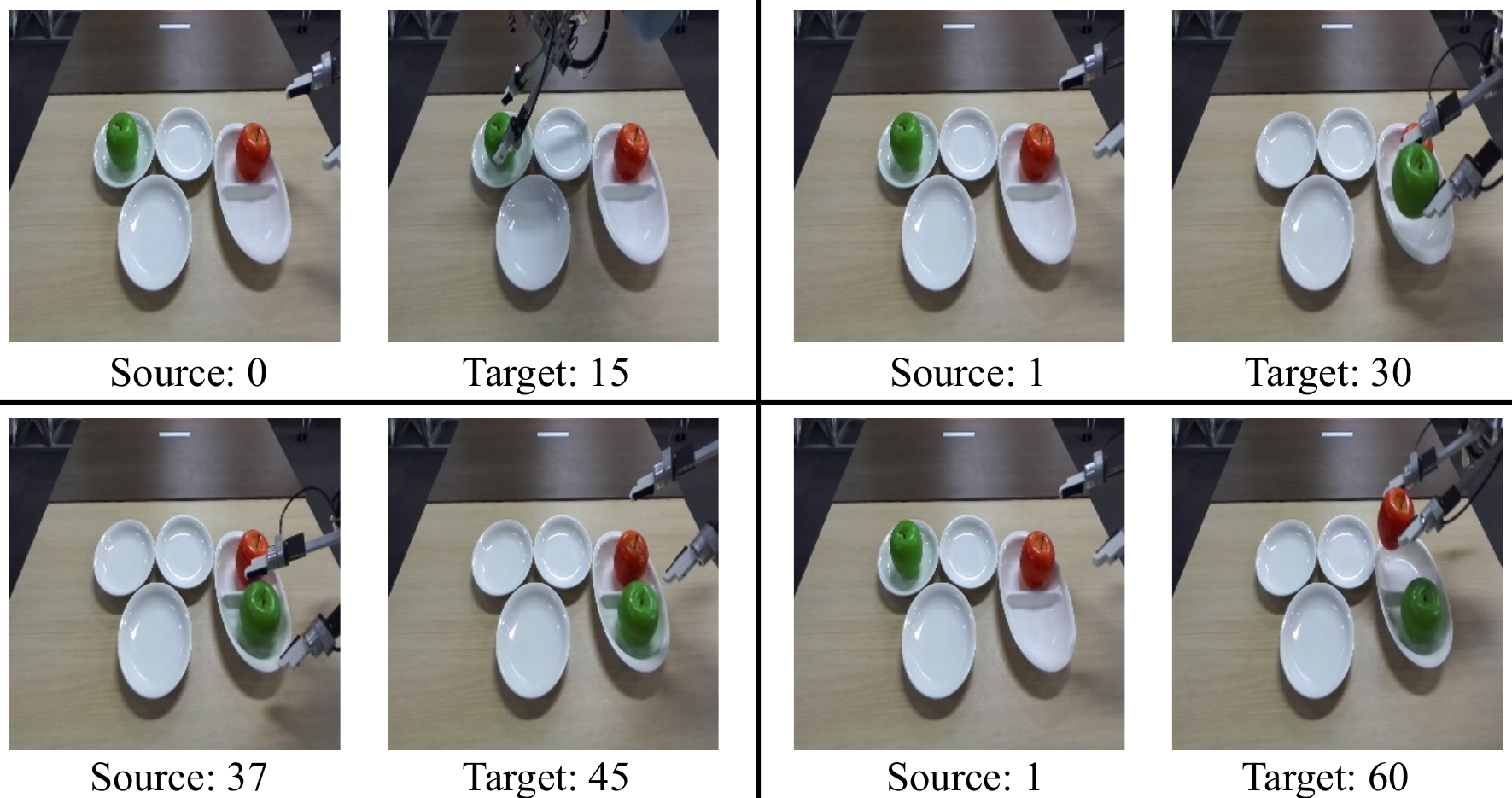}
    \caption{Sample pair images at the target sequence and the most attended source sequence.}
    \label{fig:attentded_scene}
  \end{subfigure}%
  \begin{subfigure}[t]{.25\linewidth}
    \captionsetup{width=1.2\linewidth}
    \captionsetup{justification=centering}
    \includegraphics[width=0.98\linewidth]{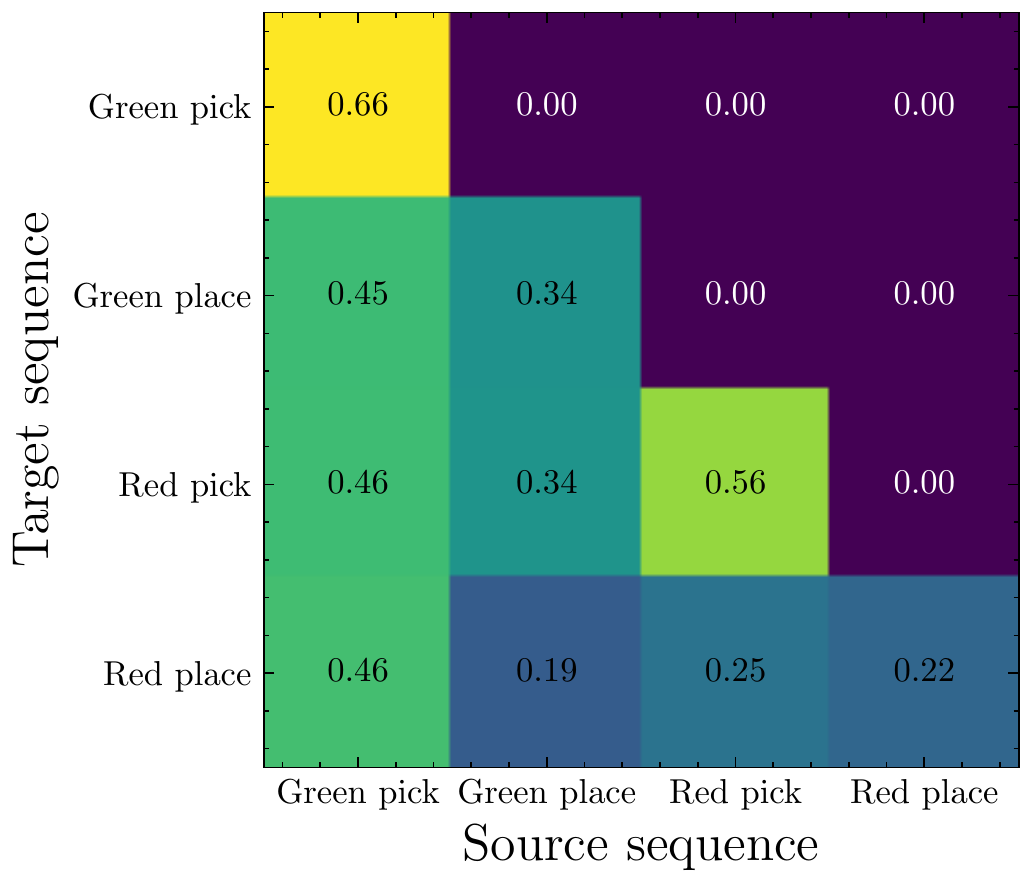}
    \caption{Averaged attention segmented by subtasks on all episodes in validation set.}
    \label{fig:hist_total}
  \end{subfigure}%

  \caption{The sequential attention map.}
  \label{fig:attention}
 \end{figure*}

The proposed Transformer-based method, LSTM baseline, and non-sequential baseline were tested on the \textit{Replace} task. The success rates for each subtask in the \textit{Replace} for 21 trials are illustrated in Fig. \ref{fig:accuracy}. In this evaluation, the proposed Transformer-based architecture recorded the highest success rate (66.7\% for the last red apple place subtask). The networks that consider sequential data were able to memorize the location of the plate where the green apple was initially placed and placed the red apple on the correct plate (Fig. \ref{fig:transformer_successful}).

In contrast, the non-sequential neural network baseline achieved a final success rate of $33.3\%$. The typical reasons for the failures of the non-sequential baseline were failure to predict the appropriate pick-and-place timing \ref{fig:ep_failure0}, or failure to gaze at the green apple's location because the network did not have any memory (Figs. \ref{fig:ep_failure1} and \ref{fig:ep_failure1_gaze}). LSTM frequently failed to pick the green apple. Also, the manipulation accuracy of the Transformer decreased without positional encoding augmentation.

\subsection{Gaze behavior analysis}

We analyzed the behavior of the gaze predictors on sequential input. 
Table \ref{tab:gaze_accuracy} (a) illustrates the successful gaze behavior of each gaze predictor model on the validation set. The successful gaze behavior is defined by properly gazing at the previous green apple plate position when the robot grasped the red apple. The Transformer-based and LSTM-based gaze predictors were able to gaze at the previous green apple plate, while the non-sequential model succeeded in only half of the trials because it does not consider any previous memories. 
Table \ref{tab:gaze_accuracy} (b) calculated the mean Euclidean distance of human gaze and the predicted gaze, which also indicates the Transformer-based method outperformed other methods.
Figure \ref{fig:gaze_comp} visualize the gaze behavior on an episode in the validation set. In step 3, while the human and the Transformer gazed at the plate where the green apple was initially located, the non-sequential model was not able to remember and find the appropriate gaze position. In step 4, the non-sequential model finally found an appropriate gaze point by tracking the robot arm's movement. However, as the robot's action is computed based on the predicted gaze position, step 4 cannot be achieved in real robot control with the non-sequential model. 

One advantage of Transformer against LSTM is that the relationship between elements in a sequence is explainable by analyzing attention weights. 
To understand the behavior of self-attention on the sequential robot data, the sequential attention map is visualized (see Appendix. \ref{sec:attention_map} for computation of the sequential attention map). Figure \ref{fig:attention_map} visualize the sequential attention map of the Transformer-based gaze predictor on one randomly selected validation episode. This indicates that while predicting the gaze between the time step of $0 \sim 30$ and $60 \sim 73$ (target sequence), the model mostly attended on the time step between $0 \sim 5$ (source sequence). However, while predicting the gaze between time step of $31 \sim 59$, the model attended on time step between $31 \sim 59$. Figure \ref{fig:attentded_scene} visualize the time step in $[0, 30, 45, 60]$ (target) and the corresponding most attended time step (source). This also indicates that while predicting gaze at time step $[15, 30, 60]$ which correspond to subtask \textit{Green pick}, \textit{Green place}, and \textit{Red place}, time step $0$ or $1$ were mostly attended. However, during gaze prediction on the subtask \textit{Red pick} which is not related to green apple's previous position, these early time steps were less attended. 

For quantitative analysis of the previous finding, Figure \ref{fig:hist_total} segmented time steps into subtasks (\textit{Green pick}, \textit{Green place}, \textit{Red pick}, and \textit{Red place}) and averaged attention values in each subtask (see Appendix. \ref{sec:attention_map} for detailed computation) on all episodes in validation set. This result indicates that gaze prediction on \textit{Red pick} mostly attended on itself, while other tasks attend on \textit{Green pick}. Especially, \textit{Red place}, which requires information of previous green plate location, attended on the time steps of \textit{Green pick}. This indicates that the Transformer-based sequential model can utilize the data on previous time steps to infer the current gaze position.

\section{Discussion}

This research investigated gaze-based deep imitation learning for sequential robot data. This approach proposed gaze prediction from sequential data to achieve robot manipulation which requires memory of previous states. First, the Transformer-based self-attention architecture for gaze prediction was proposed. Then, we investigated whether the proposed method for gaze estimation with sequential data could enable the robot to gaze at a position that can only be inferred from the memory of the past experience, therefore leading to successful manipulation. This method can be trained with human gaze estimated during the demonstration; no further annotated labels for object detectors are required.

In our experiment, LSTM-based architecture can also predict gaze position, though the Transformer-based architecture outperformed in task success rate. Empirically, the gaze prediction can be less sophisticated than policy prediction, which has to cumulatively computes the correct trajectory. Thus, this result is not surprising. However, it is difficult to train an LSTM with a sequence that consists of more than 1,000 time steps \cite{li2018independently}. 
Therefore, LSTM may not be suitable for real-world tasks that require a longer memory. Because the Transformer does not backpropagate gradients through time steps, the Transformer's self-attention is a promising approach for such tasks.

One problem while handling very long sequential data with a Transformer is the memory problem, because the Transformer's computational and memory complexities are on the order of $\bigO(n^2)$. 
To train the Transformer with very long sequential robot data, a more memory and computationally efficient architecture is needed. Recently, there have been studies on efficient Transformer training (e.g., \cite{li2019enhancing,wang2020linformer,zhou2021informer}). The adoption of such methods in robot learning may realize complex manipulation in real life.

\appendix
\subsection{Sequential attention map calculation}\label{sec:attention_map}
On each sequential input with length $L$, the Transformer computes attention weight $A^{L \times L}$. Because we want to analyze attention to the source sequence on every target time step, this attention weight is normalized by the maximum value in each target sequence:
\begin{equation}
\begin{aligned}
\label{eq:attention}
M_{i,j} = \frac{A_{i,j}}{max(A_{i})},
\end{aligned}
\end{equation}
while $M$ refers to sequential attention map and $i \in [0, L)$, $j \in [0, L)$.

Also, in all subtasks \textit{Green pick} $\in [0,s_{gp})$, \textit{Green place} $\in [s_{gp},s_{gr})$, \textit{Red pick} $\in [s_{gr},s_{rp})$, and \textit{Red place} $\in [s_{rp},L)$, where $s_{gp}, s_{gr}, s_{rp}$ indicate segmentation point at each subtask, the subtask attention map $S$ is computed by averaging all attentions in the subtask. For example, attention of target \textit{Red pick} to source \textit{Green pick} is computed by:
\begin{equation}
\begin{aligned}
\label{eq:subtask_attention}
S_{Red pick,Green pick} = \frac{1}{s_{gp}(s_{rp}-s_{gr})}\sum_{j=0}^{s_{gp}} \sum_{i=s_{gr}}^{s_{rp}} M_{i,j}
\end{aligned}
\end{equation}

\bibliographystyle{IEEEtran}
\bibliography{IEEEfull}

\end{document}